\documentclass[sigconf,nonacm]{acmart}

\usepackage{booktabs}
\usepackage{multirow}
\usepackage{graphicx}
\usepackage[utf8]{inputenc}
\usepackage[T1,T5]{fontenc}

\graphicspath{{images/}}

\title{Automated Lexical Coverage for Language Learning: From General to Specialized Word Lists}

\author{Dakota Ellis}
\authornote{Corresponding author}
\affiliation{
  \institution{University of North Carolina at Charlotte}
  \department{School of Data Science}
  \city{Charlotte}
  \state{NC}
  \country{USA}
}
\email{dellis23@charlotte.edu}

\author{Samy Babikerali}
\affiliation{
  \institution{University of North Carolina at Charlotte}
  \department{School of Data Science}
  \city{Charlotte}
  \state{NC}
  \country{USA}
}
\email{sbabike2@charlotte.edu}

\author{Wanshan Chen}
\affiliation{
  \institution{University of North Carolina at Charlotte}
  \department{School of Data Science}
  \city{Charlotte}
  \state{NC}
  \country{USA}
}
\email{wchen38@charlotte.edu}

\author{Bảo Đinh}
\affiliation{
  \institution{University of North Carolina at Charlotte}
  \department{School of Data Science}
  \city{Charlotte}
  \state{NC}
  \country{USA}
}
\email{bdinh2@charlotte.edu}

\author{Uyen Le}
\affiliation{
  \institution{University of North Carolina at Charlotte}
  \department{School of Data Science}
  \city{Charlotte}
  \state{NC}
  \country{USA}
}
\email{ple12@charlotte.edu}

\begin{document}

\begin{abstract}
A General Service List (GSL) is a commonly used resource for language learners to identify important English words. Traditional GSL creation is resource-intensive, relying
on linguistic expertise and subjective input. We created our own GSL and evaluated its performance against the New General Service List (NGSL). We found that creating a Specialized Word List (SWL), tailored to a specific text, is a practical method for language learners. Because an SWL is derived from the target text itself, it reaches the 95\% coverage required for language comprehension by construction, and it does so with substantially fewer words than a general list applied to the same text: across nine texts spanning fiction, academic papers, and scripts, the NGSL covered 64--85\% of each text, whereas a text-specific list reached 95\% with far smaller vocabularies. By restricting the SWL process to objective criteria only, it can be automated, scaled, and tailored to the needs of language-learners across the globe.
\end{abstract}


\maketitle

\section{Introduction}

Language learning is central to communication, education, and global opportunity \cite{unesco2016}. For learners worldwide, acquiring English proficiency opens doors to academic success, career advancement, and cross-cultural understanding \cite{crystal2003}. Language proficiency is commonly described in terms of four core skills: listening, speaking, reading, and writing. Among these, reading is a key determinant of fluency \cite{nation2006}. However, a principal barrier to reading comprehension is lexical coverage \cite{laufer2010}, or the proportion of words in a text that a learner knows.

Research shows that learners must recognize at least 95\% of the words in a text to achieve reasonable comprehension without external support \cite{laufer2010}. Besides reading comprehension, vocabulary knowledge is a strong predictor of academic success and listening ability \cite{bauer1993, laufer2010, laufer2020}. With insufficient vocabulary knowledge, comprehension is fragmented, cognitive load increases, and reading becomes frustrating \cite{hu2000}. In other words, adequate vocabulary knowledge is integral to language comprehension \cite{bauer1993, laufer2010}.

To address frustration of insufficient vocabulary knowledge, educators and researchers have relied on wordlists to guide vocabulary instruction \cite{vukovic2022}. These are curated sets of vocabulary extracted from large text corpora and are designed to help learners meet the comprehension threshold \cite{bauer1993}. The most influential example is West's \textit{General Service List} (GSL) \cite{west1953}. More recently,  efforts have generally revolved around addressing concerns about the GSL's age \cite{browne2013, brezina2013} and applicability to specific genres \cite{coxhead2000, konstantakis2007, wang2008}. Despite these efforts, these resources still frequently fail to achieve sufficient coverage for individual texts. They are static in nature and rely on corpus-level, aggregated frequency statistics, which can overlook the evolution of language and unique vocabulary demands of specific materials \cite{nation2006}. Therefore, learners face barriers to understanding texts across diverse domains such as legal documents, academic papers, or entertainment media \cite{nation2006}.

To examine the scalability of corpus-based wordlist construction, we attempted to update the GSL. In a similar vein to those before \cite{browne2013, brezina2013}, we used a larger, more contemporary corpus, the Google Books Ngram Corpus \cite{michel2011}. While this effort yielded a broader and more modern vocabulary, we found that increasing corpus size did not meaningfully improve the utility of the list. In fact, coverage gains were often arbitrary, driven more by decisions about list size or the method of grouping word variants than by pedagogical effectiveness. This confirmed a deeper, structural issue: even updated wordlists remain static and corpus-bound; therefore, they struggle to account for the dynamic nature of language learning.

In this project, we propose a scalable solution: a personalized wordlist generator that guarantees at least 95\% vocabulary coverage of any given text. Unlike traditional lists, our method operates directly on the input text that requires no external corpus, linguistic tools, or expert knowledge. By leveraging Zipf's law \cite{zipf1949} and frequency-based coverage calculations, we show that learners can generate focused, high-impact vocabulary lists tailored to their reading needs. This approach expands access to reading comprehension tools, empowers learners to bridge the vocabulary gap on demand, and aligns learning resources with individual needs and real-world tasks.

\section{Background}

Word lists are essential for curriculum design, assessment, and self-guided study. They aim to provide the most relevant words for learners to know in order to achieve language comprehension \cite{vukovic2022}. They are typically compiled using frequency analysis of large corpora that represent a particular target language, which is often general English. The words listed in a wordlist, called headwords, must cover a large portion of their target language and its variants must be readily recognizable to learners. However, these methods hinder their adaptability to domain-specific texts and the individual learners' goals. We review existing wordlists, how they are constructed, their shortcomings in meeting lexical coverage thresholds, introduce Zipf's law as a statistical basis for vocabulary distribution, and motivate our approach for generating high-coverage word lists efficiently and at scale.

\subsection{Existing Wordlists}

In their prominent 1953 GSL, West selected about 2,000 headwords from a 5-million-word English corpus based on manual calculations of frequency and conceptual coverage criteria \cite{west1953}. In practice, West's list covers 75\% of nonfiction texts and 90\% of novels but has since been criticized for a lack of vocabulary from specialized texts \cite{coxhead2000} and its age \cite{vukovic2022}.

To supplement West's list with domain-specific vocabulary, a wide range of Language for Specific Purposes (LSP) wordlists have since been proposed for English learners. Coxhead's well-known 2000 \textit{Academic Word List} (AWL) supplements the GSL using a 3.5-million-word academic corpus \cite{coxhead2000}. They selected 570 headwords to account for an additional 10\% of academic texts on top of GSL's coverage \cite{coxhead2000}. Aside from general academic English, more domain-specific lists have been proposed \cite{vukovic2022}. For example, Konstantakis's 2007 \textit{Business Word List} \cite{konstantakis2007}, Wang, Liang and Ge's 2008 \textit{Medical Academic Word List} \cite{wang2008}, or Minshall's 2013 \textit{Computer Science Word List} have been proposed for learners to familiarize themselves with the respective fields' jargon \cite{vukovic2022}.

More recently, Browne et al. addressed the concern about the GSL's age in their 2013 \textit{New General Service List} (NGSL). Based on a 273-million-word corpus, the NGSL selected 2809 headwords to cover 92\% of general, written English \cite{browne2013}. Since their update to the GSL, they have supplemented the NGSL with word lists specific to academic, business, and Test of English for International Communication (TOEIC) English.

\subsection{Wordlist Construction is Resource-Intensive}

In all cases, the construction of high-quality wordlists was resource intensive. Three key challenges underpin this process: the sourcing of large, representative corpora, the need for domain and language acquisition expertise, and the reliance on specialized lexicographic tools often inaccessible to educators and learners.

The foundation of any credible wordlist is the corpus it was derived from \cite{vukovic2022}. The corpus must be representative of the target language. For example, Coxhead's academic corpus gathered texts from four disciplines and multiple registers \cite{coxhead2000}. Similarly, the NGSL derived its words from a subset of the Cambridge English Corpus \cite{browne2013}. Instead of using the entire corpus, Browne et al. omitted the Cambridge English Corpus's newspaper and academic sub-corpora because they biased the corpus away from general English \cite{browne2013}. Furthermore, access to large, balanced corpora remains a barrier for many. Such corpora often demand institutional access to proprietary databases needing legal permission to redistribute or repurpose the texts. This problem presents a significant obstacle for individual researchers, educators, and learners \cite{davies2009}. Indeed, the challenge of sourcing accessible corpora significantly shaped our motivation to conduct this research.

Once the corpus is compiled, selecting headwords is not merely a mechanical process. After validating the contents and balance of the corpus, the lexicographer must make decisions about headword representation and selection criteria \cite{brezina2013, browne2013}. For headword representation, the method of grouping variants into a singular, representative headword often follows Bauer and Nation's 1993 framework \cite{bauer1993}. As mentioned before, the grouping method is chosen at the discretion of the lexicographer under the assumption that learners will be able to recognize variants once the headword is known \cite{bauer1993}. For example, both West \cite{west1953} and Coxhead \cite{coxhead2000} grouped words into word families. In their case, the derivational and nominal forms of develop (verb), such as undeveloped and development, fall within the scope of develop. However, the ease of recognizing variants when grouped under word families has been criticized \cite{brezina2013}. In recent decades, lemmas are the preferred grouping of recent wordlists \cite{brezina2013, browne2013}. Besides frequency, headwords may be selected based on dispersion, keyness, or domain-relevance at the discretion of the lexicographer \cite{vukovic2022}. For instance, West used pedagogical judgment to exclude words deemed too technical or narrow in use \cite{west1953}. The reliance on expert evaluation introduces subjectivity and limits scalability. Few linguists have the necessary background to conduct work rigorously \cite{cobb1995}.

Even when qualifying corpora are available, processing them into usable wordlists requires specialized tools: concordancers, frequency analyzers, lemmatizers, and corpus query systems. These tools often require licenses and come with a steep learning curve. SketchEngine, for example, is a powerful corpus query tool used by many lexicographers \cite{browne2013}. However, it is a commercial product with subscription fees and is designed for lexicographers. AntConc is free but assumes familiarity with corpus analysis methods and isn't user-friendly. For language instructors and learners, these tools are either too complex or out of reach, further entrenching the gap between corpus research and classroom practice.
The dependency on expert opinion and inaccessible resources contributes to why most existing wordlists are static. This results in a reliance on outdated or overly general lists that fail to meet learners' comprehension needs in specific contexts. It is precisely this inefficiency and inaccessibility that motivates the development of dynamic, text-specific wordlist generation.

\subsection{Reading Comprehension Thresholds}

Research on vocabulary acquisition has identified that learners must know 95\% to 98\% of the words in a text to reasonably understand it \cite{hu2000, laufer2010}. Within the context of reading comprehension, "reasonable" is a probabilistic threshold where learners above this threshold have a chance to understand the text \cite{laufer2010}. In other words, most learners but not all will have marginal understanding at 95\% coverage and must reach 98\% for adequate understanding \cite{hu2000, laufer2010}. In practice, learners must know 4,000 to 5,000 word families to achieve 95\% of general English \cite{nation2006} which few learners are able to achieve without guidance \cite{laufer2010}. While 98\% is a safer pedagogical target to ensure learners' comprehension, 8,000 to 9,000 word families are required. The key insight is that lexical coverage relates to comprehension non-linearly. Going from 95\% to 98\% coverage requires nearly the same number of word families to be known as going from 0\% to 95\% coverage. In terms of wordlists, the list must enlarge dramatically to achieve marginal increments in coverage.

Furthermore, wordlists often fail to come close to these thresholds in authentic texts making the task of reaching reading comprehension even harder. For example, Coxhead's AWL was born out of the GSL covering only 75\% of academic corpora \cite{coxhead2000}. Even when combined together, the GSL and AWL account for just 90\% of a 600,000-word business corpus with specialized, frequent terms missing \cite{konstantakis2007}. Similar analyses for medical terms show that the AWL covered only a fraction of the vocabulary \cite{wang2008}. In short, no published list provides 95\% coverage for all text types. Even large corpora-derived lists like the NGSL fall short of 95\% for its target language of general English \cite{browne2013}. Moreover, static lists cannot account for text-specific proper nouns, formulas, or recent coinages. As a result, when learners encounter authentic materials, coverage with existing lists often falls short of necessary levels.

\subsection{Learner Frustration at Low Coverage}

When lexical coverage is too low, learners experience immediate comprehension breakdown and frustration. At 98\% coverage, learners will see a new word every five lines; for lower coverages, learners will experience frustration every two lines for 95\% coverage and every single line for 90\% coverage \cite{hu2000}. While good inference ability can compensate for lower coverage \cite{laufer2020}, unfamiliar words disrupt the flow of reading and require repeated dictionary use \cite{hu2000}. Such interruptions result in incomplete comprehension and a feeling of stalled progress \cite{hu2000}. In other words, texts below 95\% coverage become "frustration-level" reading.

With learners' frustration in mind, the relevance of text-specific wordlists to learners becomes clear, particularly for learners studying independently. Indeed, four of our members are non-native speakers who are intimately familiar with English as a Second Language (ESL) pedagogy and rely on previewing vocabulary before reading. As such, the usefulness to language learners was a motivating factor in our decision to pursue this research.

\subsection{Purpose of this Study}

The purpose of this study is to evaluate the effectiveness of GSLs by creating our own method and comparing its performance with the industry standard. We also propose a bottom-up, text-specific approach to wordlists. Instead of relying on one-size-fits-all lists, this method derives a \textit{Specialized Word List} (SWL) directly from the target text's own frequency profile. First, an automated analysis ranks the text's words by occurrence and includes the most frequent items in the learner's list. Since the list is tailored to the text, the method guarantees high coverage. Crucially, this strategy aligns with Zipf's law, which states that word frequencies in any natural text follow a power-law distribution \cite{zipf1949}. In other words, a few extremely frequent words cover a large portion of the text. By exploiting this principle, a high-coverage, text-specific wordlist can be generated for virtually any text given that it follows a power-law distribution. 

The text-specific list adapts to the text's domain, eliminating irrelevant items and addressing unfamiliar words. Learners can focus on acquiring the unfamiliar vocabulary they need to achieve the comprehension threshold. This targeted preparation enables more immediate and tangible progress, boosting learners' confidence and motivation \cite{hu2000}. When learners feel capable, they are more likely to continue their efforts \cite{bandura1982}.

\section{Methodology}

\subsection{The Google Books Corpus}

The Google Books Corpus \cite{michel2011} is a dataset containing the word frequencies from over 5 million books gathered from the year 1500 through 2022. The full corpus represents 6\% of all books ever published at the time of its last update. Only words that occur 40 or more times within their respective corpus are added to the Google Books dataset. This means that unique, low-frequency  words will not appear within this corpus. The records in the dataset contain the word token, the year in which it was used, the total number of times the word appeared, and the count of books that the word appeared in. The English section within the corpus contains 500 billion words. We preprocessed the subset used for our GSL to include two main features, \textit{Count} and \textit{Word}. \textit{Count} is the word frequencies contained in the books compiled by Google's Ngram. This was done by first filtering out any years before 1800 and grouping the dataset by \textit{Word}, which generated the frequency word list of 9 million words ranging from 1800 to 2019. The \textit{Word} feature is where most of the pre-processing took place. The Ngram included numbers, symbols, corrupted tokens (e.g., Ã, â€"), compounded words (e.g., state-of-the-art), possessives, and contractions. All of these strings have been filtered out and removed using regular expressions. This step entailed case normalization to avoid capitalization variation and ensure that each word is unique in the list. However, at this step some impurities still remain, such as misspelled words, and scanning errors, which will be addressed later in the analysis.

\begin{figure*}[t]
  \centering
  \includegraphics[width=\textwidth]{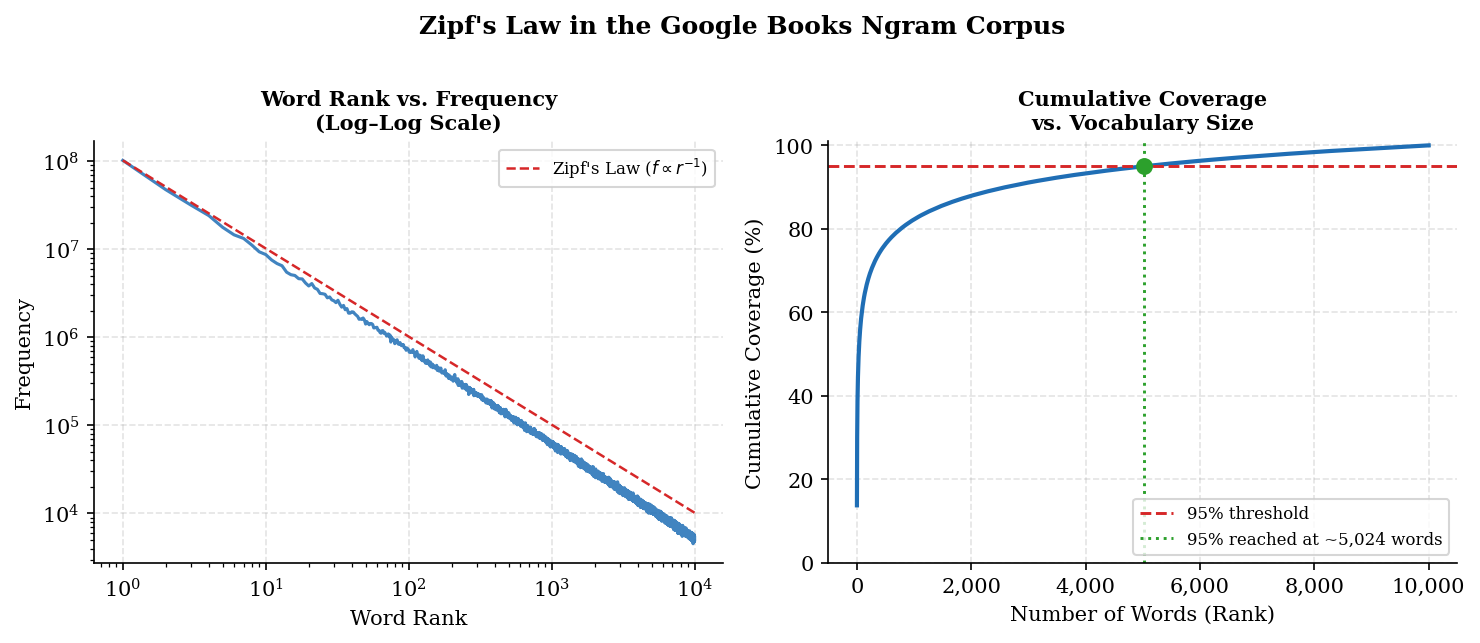}
  \caption{Zipf's Law in the Google Books Ngram corpus. \textit{Left}: log--log plot of word rank vs.\ frequency; the near-linear fit (dashed) confirms a power-law distribution. \textit{Right}: cumulative coverage curve showing that 95\% coverage is reached at approximately 5,000 words, motivating the use of a frequency-ranked cutoff for wordlist construction.}
  \label{fig:histogram}
\end{figure*}

\subsection{Calculating Coverage}

A word's coverage refers to the percentage of the corpus that comprises that word. For example, if a book contains 1000 total words or tokens and the word \textit{dog} appears 20 times, then \textit{dog} has a coverage of 2\%. It is important to note that coverage can be accumulated through addition. If \textit{cat} has a coverage of 1\% in that same book, then a word list made up of two words \textit{dog} and \textit{cat} has a coverage of 3\% for that book.

\subsection{Creating the GSL}

When we created our GSL, the goal was to apply traditional methodologies to a large corpus of our choosing. The corpus needed for creating a GSL must be chosen carefully to ensure that it is representative of the language as a whole and must contain several examples of literature of multiple genres. We chose to utilize a large subset of the Google Books Corpus, containing the top 9 million words since the year 1800. Once we felt confident that our corpus was sufficient in representing the English language, we started by tokenizing the corpus into unigrams (1-grams). We then calculated the frequency of each unique word by counting how many times each word appears within the corpus. After this, we reduced the word list to headwords by applying lemmatization, grouping words into base form by word family. After ranking the headwords by frequency, we plotted the cumulative coverage gains against their rank positions. The derivative was then taken to find the point of diminishing return for coverage gained by each increment in rank. This yielded a value \textit{p} that corresponded to the rank in which all headwords from rank 1 to rank \textit{p} would be the highest frequency words within the corpus. These words were selected for the GSL.

\begin{figure}[t]
  \centering
  \includegraphics[width=\linewidth]{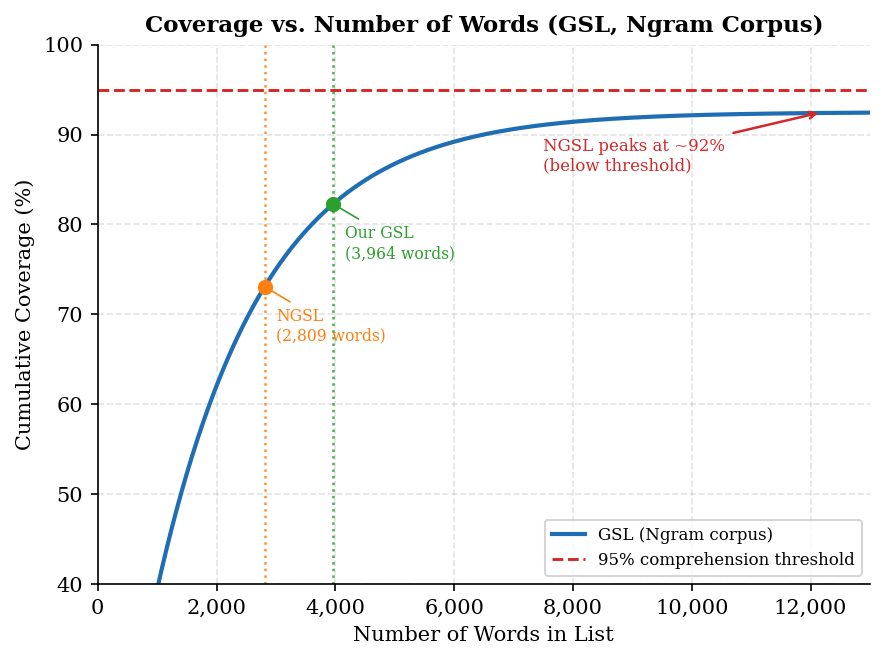}
  \caption{Cumulative lexical coverage as a function of word-list size in the Google Books Ngram corpus. The 95\% comprehension threshold (dashed red) is never reached by either the NGSL (2,809 words, orange) or our GSL (3,964 words, green), illustrating the structural limitation of general-purpose lists on specific texts.}
  \label{fig:coverage_vs_words}
\end{figure}

\subsection{Creating the SWL}

Constructing an SWL follows a similar process to GSL's with a few key differences. The corpus does not need to be representative of the language as a whole, and the optimal word list is determined based on the 95\% coverage requirement. For our SWL's, we used fiction novels, academic papers, and TV show and movie scripts. These corpora represent common forms of English that language learners consume. Once each corpus was tokenized, counted, and lemmatized, the headwords were sorted in rank order. The significant words were those from rank 1 to rank \textit{p}, where cumulative coverage of rank \textit{p} is equal to 95\%.

Our model will be tested on three different texts: a textbook (\textit{Text Data Management and Analysis}), a novel (\textit{Alice in Wonderland}), and a movie script (\textit{Titanic}). This evaluation method tests the model in three different genres to highlight its versatility and applicability across any field.

The model was designed using the Python programming language, as Python is widely used for processing data, including text. Additionally, many natural language processing (NLP) libraries are created in Python, making it an ideal tool for this model. The libraries utilized in this model are NLTK, SpaCy, and Stanza: Natural Language Toolkit (NLTK) is responsible for tokenizing and removing stop words; SpaCy is used for lemmatization; and Stanza is used to remove proper nouns from the text.

\begin{figure*}[t]
  \centering
  \includegraphics[width=\textwidth]{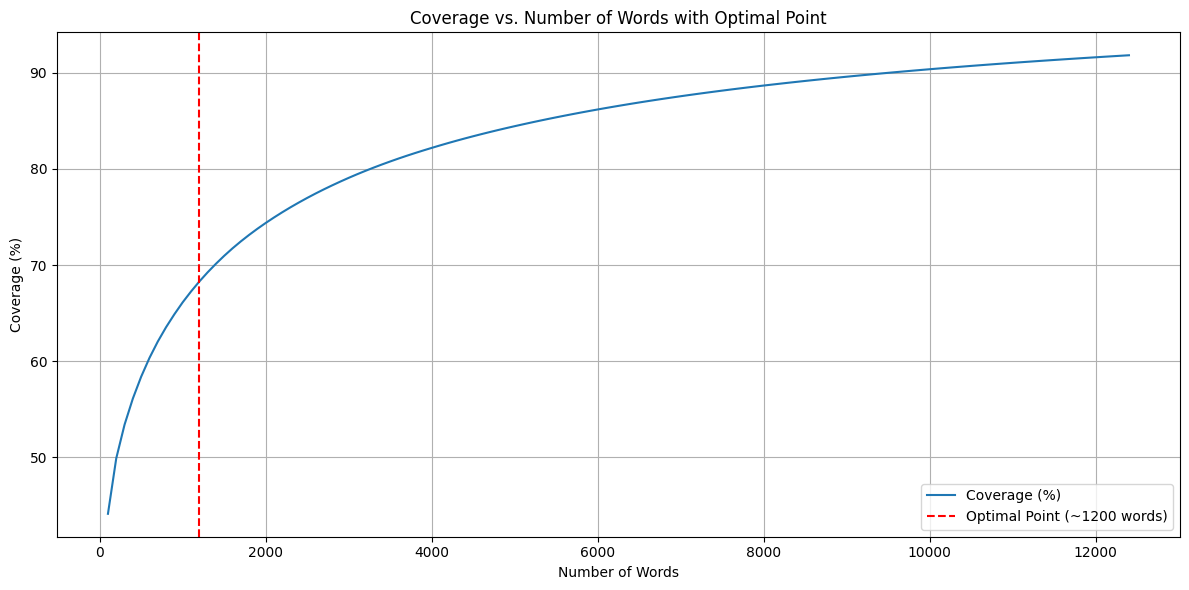}
  \caption{Coverage versus word-list size with the diminishing-returns inflection marked at $\sim$1,200 words. The rapid early gains and subsequent flattening illustrate the Zipfian behavior that motivates the frequency-ranked cutoff used in our SWL methodology. As discussed in Section~\ref{sec:gsl-dev}, this inflection did not by itself yield a pedagogically adequate GSL cutoff, which is why our final SWL criterion fixes coverage at 95\% rather than relying on this turning point.}
  \label{fig:coverage_optimal}
\end{figure*}

\section{Results}

\begin{tabular}{lrrrr}
\toprule
\multirow{2}{*}{Literature} & \multicolumn{2}{c}{NGSL} & \multicolumn{2}{c}{SWL} \\
\cmidrule(lr){2-3} \cmidrule(lr){4-5}
& Coverage & Size & Coverage & Size \\
\midrule
  Alice in Wonderland & 88.8\%   & 2809 & 95\%  & 664 \\
  Pride and Prejudice & 85.2\% & 2809 & 95\%  & 1418 \\
  The Great Gatsby    & 81.6\% & 2809 & 95\%  & 2077 \\
\midrule
  Image Recognition   & 64\%   & 3766 & 95\%  & 845  \\
  AI and Proteins     & 65\%   & 3766 & 95\%  & 887  \\
  Neural Networks     & 66\%   & 3766 & 95\%  & 947  \\
\midrule
  Text Data Mgmt & 80.1\% & 3766 & 95\%  & 1423 \\
\midrule
  Titanic             & 79.4\% & 2809 & 95\%  & 2049 \\
  Snakes on a Plane   & 76.2\% & 2809 & 95\%  & 224  \\
\bottomrule
\end{tabular}

\subsection{Our GSL}

To evaluate our GSL against the NGSL, we computed the coverage of the words in the NGSL based on the frequency count of Google's Ngram. The NGSL achieved 62\% coverage in the Ngram with a list size of 2809. Our GSL achieved a coverage of 78.4\% with a list size of 2809; The coverage grew to 82\% when the size was increased to 3964.

\begin{figure}[t]
  \centering
  \includegraphics[width=\linewidth]{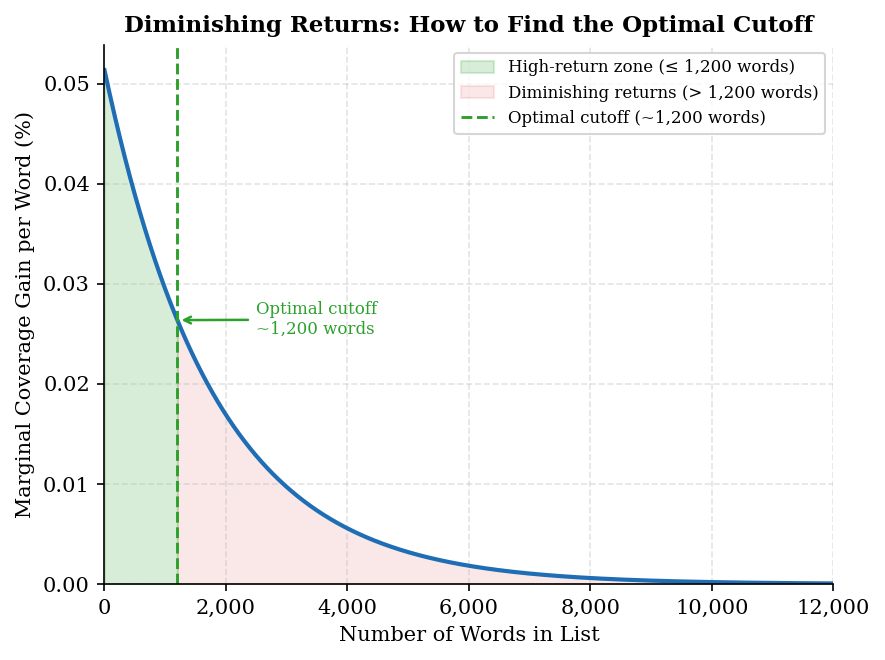}
  \caption{Marginal coverage gain per additional word in the Ngram corpus. The green shaded region marks the high-return zone (up to $\sim$1,200 words) where each new headword contributes meaningfully to coverage; beyond this point returns diminish sharply, identifying the practical GSL cutoff~\textit{p}.}
  \label{fig:threshold_vs_words}
\end{figure}

\subsection{Text Data Management and Analysis (SWL)}

The textbook used is \textit{Text Data Management and Analysis} by Dr. ChengXiang Zhai. Following the model order of steps, the textbook is loaded, tokenized, and processed to generate a frequency list. This results in 6330 unique words, which are then further reduced by lemmatizing to 4621. The next step is to compute the cumulative frequency and remove words with coverage greater than 95\%. This step generates a word list with 1423 words.

The evaluation of this list results in 95\% coverage in the original list, while the coverage of the NGSL results in 80.13\% in the original list. 

\subsection{Alice in Wonderland (SWL)}

The novel used is \textit{Alice in Wonderland} by Lewis Carroll. Similarly, using the model to generate a frequency list results in 1432 words; lemmatizing the frequency list reduces the list to 1131 words. Finally, generating the word list by removing words with cumulative coverage greater than 95\% results in 664 words.

Again, the evaluation of the word list results in 95\% coverage in the original list, while the coverage of the NGSL results in 88.75\%. It is important to also note that the word list contains only 664 words compared to the NGSL's 2809 words, while still resulting in greater coverage.

\subsection{Titanic (Movie Script) (SWL)}

Finally, to generate a word list for the \textit{Titanic} movie script, generating the frequency lists results in a list of 5341 words; lemmatizing reduces the list by 1291 words, resulting in a list of 4050 words. Removing the words with coverage greater than 95\% results in 2049 words.

The evaluation of this word list results in 95\% coverage in the original list, while the coverage of the NGSL results in 79.36\%.

As highlighted in the results above, for the task of preparing to read one specific text, an SWL reached the 95\% comprehension threshold with a far smaller vocabulary than the NGSL achieved on the same text. We note that the SWL attains exactly 95\% by construction, since that coverage level is its stopping criterion; the substantive comparison is therefore not whether the SWL reaches 95\% (it does so by design) but how few words it requires to do so relative to a general list applied to the same material. On every text tested, the text-specific list was both smaller and higher-coverage than the NGSL on that text.

\section{Discussion}

\subsection{Development and Evaluation of Our GSL}
\label{sec:gsl-dev}

Our initial goal of creating our GSL was to objectively identify an optimal threshold within the distribution where the most significant words could be found. It became difficult to identify an objective turning point where the coverage gained was not proportional to the increase in the size of the word list, \textit{n}. The initial methods for determining this point included taking the derivative of the coverage gained with each increase in rank. However, this calculation does not produce any meaningful result as the coverage of our GSL did not sufficiently prepare a language learner for literature comprehension. Additionally, it is impractical to have a language learner attempt to utilize a GSL that would have been derived from 95\% coverage of the Ngram corpus as the amount of words in the list would be in the tens-of-thousands. Logically, we determined that 95\% coverage of a specific word list for a piece of literature would meet the comprehension threshold and the size of the list would be practical for learning. This could be useful for language learners trying to understand legal documents, academic papers, or entertainment media.

\subsection{GSL vs SWL}

A GSL such as the NGSL is a list of the most frequently occurring headwords from an entire language. In contrast, an SWL is a list of the most frequently occurring headwords in a specific genre or article of language, such as a book, tv show, movie, or conversation of spoken dialogue. When comparing the effectiveness of a GSL on specific articles of language, it becomes apparent that coverage usually falls well below the 95\% coverage required for comprehension. An SWL such as the lists that we created will often out-perform a GSL in coverage and with fewer words comparatively. An SWL will also consider words created by an author within its analysis while a GSL will not, which may be of more importance when considering fantasy stories such as \textit{The Lord of the Rings} or even academic papers where new terms are coined. We see that the SWL's achieved higher coverage for the \textit{Image Recognition, AI and Protein, and Neural Network} academic papers, even when the NGSL was equipped with the Academic Word List.

\subsection{Zipf's Law Applies to Most Cases}

Zipf's Law states that the frequency of a word will be inversely proportional to its rank. This means that the coverage gained for $n+1$ where $n$ is the word list rank, will see diminishing returns suggesting that the shortest way to achieve the 95\% coverage is by selecting the headwords in rank order. The empirical evidence that shows that Zipf's law represents most domains of English supports that our research will apply to nearly all instances of an English corpus. Since the methods applied to the corpus to derive the word list are purely quantitative and require no subjective criteria, the process can be automated by an algorithm. The combination of automation and versatility means that our method makes the process of creating a specialized word list scalable to modern needs. Generating word lists will no longer require years of work or linguistic expertise, providing access to word lists to English learners of any background. By creating a word list for a specific piece of literature, the word count for 95\% coverage becomes much more practical for the user to learn.

\subsection{Limitations}

Some important aspects of our model to note would be the reliance on Zipf's law. While Zipf's law is representative of nearly all examples of natural English, there may be scenarios when a corpus deviates from a power law distribution. For instance, if a poem is written in such a way that the frequency of the words follows a uniform distribution. There is also the limitation that not all languages will follow Zipf's Law; therefore, this methodology is specific only to the languages that do. Another limitation of the SWLs that we created is that they are only frequency based and do not account for information gained by each word. This could lead to loss of in-context significance.

Two further limitations bear on how our comparison should be read. First, each SWL is both derived from and evaluated on the same text, so its 95\% coverage is guaranteed by the construction procedure rather than demonstrated out of sample; a stronger test would generate the list from one portion of a text or corpus and measure coverage on a held-out portion, which we leave to future work. Second, a general list such as the NGSL is intended to be learned once and reused across all texts a learner encounters, whereas an SWL serves a single text. The fairer comparison for a learner reading many texts is therefore between the NGSL and the \emph{union} of the SWLs required, accounting for vocabulary shared across texts. Our results establish the efficiency of an SWL for previewing an individual text, not that text-specific lists are more economical than a general list over a learner's entire reading load.

\subsection{Future Work}

By creating an app that allows language learners to create their SWLs that are specific to their needs; features that facilitate in-context learning, pronunciations, and definitions could add depth to the learning process. A dynamic library of known words specific to the user could further optimize the process of learning significant words, ensuring that they save time by avoiding unnecessary redundancy. Suggesting similar literature or media to the language learner based on their known vocabulary could strengthen the confidence of the language learner's abilities. Also, it would be crucial to include audio of spoken language to allow the language learner to understand pronunciation and vernacular. Further research should be done to find if lemmatizing before reducing to the 95\% coverage word list yields the same or very similar results as lemmatizing after reducing the corpus. If both methods produce the same list of significant headwords, then there may be some computational relief by lemmatizing later in the methodology. This may not make a meaningful difference in the computational time complexity of producing an SWL for a textbook but should make a significant difference in the time required to create an SWL for an extremely large corpus, such as the Google Ngram Books Corpus.

\subsection{The Need for Modularity}

To meet the needs of the end user, the process of creating the word list needs to be modular. For instance, as we developed our word list for \textit{Titanic}, we encountered an issue where traditional methods negatively impacted the list's utility. The word \textit{Rose} appears near the top of the frequency distribution. When this word is reduced to its lemma (as per traditional methods) it gets treated as the past tense of \textit{Rise}. This is an issue because \textit{Rose} is the name of the main character and lemmatizing will improperly add the word \textit{Rise} to the word list. While proper nouns get dropped from traditional methodologies, we felt that it is important for a language-learner to recognize names of characters as words that commonly appear as names. For instance, a native English speaker may see the name \textit{Cameron} and understand that it is a common name that could apply to multiple people, whereas an English learner may not immediately make that connection. The model that we constructed will take parameters specified by the user to ensure its applicability to specific needs. The user will be able to specify if they want to lemmatize their list, remove stop words, adjust coverage score, remove proper nouns, or limit the word list to a specific size.

\section{Conclusions}

As we have established in our demonstrations, applying a GSL to a much smaller subset of the language negatively impacts its coverage performance, often falling well below the 95\% coverage required for language comprehension. By attempting to be a representation of the entire corpus of English, a GSL does not effectively represent high-frequency words in all subsets of the language. This research highlights the benefits of a quick, automated process that can be applied to a specific piece of English literature to account for this limitation in traditional methods. We found that applying this framework to the challenge of language learning not only produced the 95\% coverage requirement, but achieved this threshold with fewer words comparatively. The creation of an SWL rather than a GSL appears to be a more efficient way for language learners to navigate the variability and variety of language.

\bibliographystyle{ACM-Reference-Format}
\bibliography{sources}
\end{document}